\documentclass{article}
\usepackage{spconf,amsmath,graphicx}
\usepackage{float,array}
\usepackage{rotating,balance}
\usepackage{relsize,cite}
\newcommand{\cb}[1]{{\boldsymbol{#1}}}
\newcommand{\cp}[1]{\ifmmode {\mathcal{#1}}\else ${\mathcal{#1}}$\fi}
\newcommand{\balpha}{\boldsymbol{\alpha}}
\newcommand{\bbeta}{\boldsymbol{\beta}}

\newcommand{\bpsi}{\boldsymbol{\psi}}

\newcommand{\bm}{\boldsymbol{m}}

\newcommand{\be}{\boldsymbol{e}}

\newcommand{\br}{\boldsymbol{r}}
\newcommand{\bv}{\boldsymbol{v}}

\newcommand{\MM}{\boldsymbol{M}}

\newcommand{\KK}{\boldsymbol{K}}

\newcommand{\bD}{\boldsymbol{D}}
\newcommand{\bI}{\mathbf{I}}

\newcommand{\UU}{\boldsymbol{U}}
\newcommand{\VV}{\boldsymbol{V}}
\newcommand{\HH}{\boldsymbol{H}}
\newcommand{\mr}[1]{{\bm_{\lambda_{#1}}}}

\newcommand{\bgamma}{\boldsymbol{\gamma}}

\newcommand{\A}{\boldsymbol{A}}

\usepackage{color}  % To include comments in color
\definecolor{olive}{rgb}{0.3,0.45,0.2}

\def\R{\ensuremath{{\mathrm{I\!R}}}}

\newcommand*{\Scale}[2][4]{\scalebox{#1}{$#2$}}%

\newcommand{\ssum}[2]{\Scale[1.01]{\sum\limits_{\mathsmaller{#1}}^{\mathsmaller{#2}}}}
\title{Nonlinear unmixing of hyperspectral images  using a semiparametric model and spatial regularization}
\name{Jie Chen$^{\,\star}$, \; C\'edric Richard$^{\,\star}$, \; Alfred O. Hero III$^{\,\dagger}$\thanks{ {This work was supported by the Agence Nationale pour la Recherche, France, (Hypanema project, ANR-12-BS03-003).}}}
\address{\small ${^\star}$ Universit\'e de Nice Sophia-Antipolis, CNRS, France \\
\small $^\dagger$University of Michigan, Ann Arbor, USA \\
\small E-mail:\quad \{jie.chen, cedric.richard\}@unice.fr,\; hero@eecs.umich.edu \vspace{-3mm}}
\vspace{-6mm}
\begin{document}
\ninept

\maketitle
\begin{abstract}
Incorporating spatial information into hyperspectral unmixing procedures has been shown to have positive effects, due to the inherent spatial-spectral duality in hyperspectral scenes. Current research works that consider spatial information are mainly focused on the linear mixing model. In this paper, we investigate a variational approach to incorporating spatial correlation into a nonlinear unmixing procedure. A nonlinear algorithm operating in reproducing kernel Hilbert spaces, associated with an $\ell_1$ local variation norm as the spatial regularizer, is derived. Experimental results, with both synthetic and real data, illustrate the effectiveness of the proposed scheme.
\end{abstract}
\vspace{-2mm}
\section{Introduction}
\label{sec:intro}
\vspace{-2mm}

Hyperspectral imaging provides two dimensional spatial images over many contiguous spectral bands. The high spectral resolution allows a comprehensive and quantitive analysis of materials in remotely observed data. This area has received considerable attention in the last decade, see~\cite{Bioucas2013Mag} for a survey.

Usually, observed reflectance at each pixel is a spectral mixture of several material signatures, called endmembers, due to limited spatial resolution of observation devices and diversity of materials. Consequently, spectral unmixing has become an important issue for hyperspectral data processing~\cite{Keshava2002}. There have been significant efforts during the past decade to address the linear unmixing problem for hyperspectral data~\cite{Heinz2001,Iordache2010,bioucas2012}. Nevertheless, the linear model can only capture simple interactions between elements, e.g., in situations where the mixing of materials is not intimate and multiple scattering effects are negligible~\cite{Keshava2002,Bioucas2013Mag}. Recently, several researchers have begun exploring nonlinear unmixing techniques. In~\cite{Halimi2011}, nonlinear unmixing was proposed based on the bilinear model and Bayesian inference.  Post-nonlinear mixing models were discussed in~\cite{Altmann2012,Chen2013-Whisp}. Unmixing algorithms using geodesic distances and other manifold learning based techniques were investigated in~\cite{Heylen2011,Nguyen2012,nguyen2013supervised,honeine2009solving}. In addition, algorithms operating in reproducing kernel Hilbert spaces~(RKHS) have been proposed for hyperspectral unmixing. Nonlinear unmixing with intuitive kernels was investigated in~\cite{Broadwater2007}. Physically-inspired kernel-based models were introduced in~\cite{Chen2011-Whisp}, where each mixed pixel is modeled by a linear mixture of endmember spectra, coupled with an additive nonlinear interaction term to model nonlinear effects of photon interactions. In~\cite{Chen2013-TSP,Chen2012-Whisp,Chen2013-Icassp},  a more complete and sophisticated theory related to this strategy was presented. See~\cite{Dobigeon2013Mag} for an overview of recent advances in nonlinear unmixing modeling.

Beyond simply providing rich spectral information, remotely sensed data convey information about the spatial variability of spectral content in the 2D terrain~\cite{Plaza2011}. Subsequently, hyperspectral analysis techniques should benefit from the inherent spatial-spectral duality in hyperspectral scenes. Following this idea, researchers have attempted to exploit spatial information for hyperspectral image unmixing. An NMF problem regularized with the $\ell_1$-norm of differences between neighboring pixels was introduced in~\cite{Zymnis2007}.  In \cite{Eches2011}, a Markov random field was proposed to model the spatial structure underlying pixels within classes. In~\cite{Mittelman2012}, unmixing by a multi-resolution sticky Hierarchical Dirichlet Process model was used to account for spatial correlations.  In~\cite{Iordache2012}, total variation was used for spatial regularization in order to enhance unmixing performance. Some other works also showed that incorporating spatial information can have a positive effect on unmixing processes~\cite{Jia2007,Zare2011,Martin2011}. Nevertheless, all these works were conducted with a linear mixing model. Rarely, if ever, have nonlinear mixing models incorporating spatial information been considered in the literature. As nonlinear unmixing is already an important but challenging issue, it appears complicated to address these two problems simultaneously. Based on the promising results of nonlinear unmixing in RKHS~\cite{Chen2013-TSP}, in this paper, we propose a novel way to incorporate spatial information into the nonlinear unmixing process using $\ell_1$-norm spatial regularization, i.e., a local version of the total variation smoothness regularizer in image reconstruction. An optimization method based on split-Bregman iterations is proposed to deal with  the nonlinearity of the model and the non-smoothness of the regularizer. 

\section{Formulation of the Problem}
\vspace{-2mm}
Consider an hyperspectral image with $w$ pixels in each row, and $h$ pixels in each column. Each pixel consists of a reflectance vector in $L$ contiguous spectral bands. In order to keep the presentation simple, we transform this three dimensional image into an $L\times N$ matrix, with $N = w \times h$ the total number of pixels. Let $n \in \{1, \dots, N\}$ be the sequential index of pixels. Suppose that the scene consists of $R$ significant endmembers, each with a spectral signature $\bm_i\in\R^ {L}$. Let $\br_n \in \R^ {L}$ be an observed hyperspectral pixel, and let $\balpha_n\in\R^R$ be the vector of endmember abundances in the pixel $\br_n$. The matrix composed of all the abundance vectors is denoted by $\A = [\balpha_1,\dots, \balpha_N]$.  Let $\MM = [\bm_1, \dots, \bm_R] \in \R^{L\times R}$  be the matrix of the endmember spectra.  For the sake of convenience, the $\ell$-th row of $\MM$ is denoted by $\bm^\top_{\lambda_\ell} \in \R^{L}$, that is, $\bm_{\lambda_\ell}$ is the vector of the endmember signatures at the $\ell$-th wavelength band. Finally, let $\cb{1}$ and $\bI$ be the all-one vector and the identity matrix, respectively, with compatible sizes according to the context.

Similarly to many nonlinear unmixing approaches, we suppose that the material signatures in the scene have been determined by some endmember extraction algorithm. The unmixing problem boils down to estimating the abundance vectors. To take the spatial relationship between pixels into consideration, the unmixing problem can be solved by minimizing a general cost function, with respect to $\A$, of the form
\begin{equation}
       \label{eq:cost.all}
       J(\A)  = J_{\text{err}}(\A) + \eta\, J_{\text{sp}}(\A)
\end{equation}
subject to a non-negativity constraint on each entry of $\A$, and a sum-to-one constraint on each column of $\A$, namely, on each $\balpha_n$. For ease of notation, these two physical constraints will be expressed by
\vspace{-2mm}
\begin{equation}
      \label{eq:constraints}
      \begin{split}
      \A \succeq \cb{0}  \qquad \text{and} \qquad \A^\top\cb{1}_R = \cb{1}_N
      \end{split}
      \vspace{-2mm}
\end{equation}
Recent work has raised the question of relaxing the sum-to-one constraint. The proposed algorithm can be easily adapted if this constraint is removed. In the experimental section, results subject to the non-negativity constraint will only be presented. In the general expression \eqref{eq:cost.all}, the function $J_{\text{err}}$ represents the modeling error and $J_{\text{sp}}$ is a regularization term to promote similarity of the fractional abundances of neighboring pixels.  Various regularizers have been defined in the literature~\cite{Zymnis2007,Iordache2012,Zare2011}. The non-negative parameter $\eta$ controls the trade-off between local data fidelity and pixel similarity.

Let us now present $J_{\text{err}}$ and $J_{\text{sp}}$ investigated in this paper. Consider the general unmixing process, acting between the entries $r_{n,\ell}$ of the observed reflectance vector, and the spectral signatures $\mr{\ell}$ of the endmembers at each wavelength band $\lambda_\ell$, defined as
          \vspace{-1mm}
\begin{equation*}
	r_{n,\ell} = \psi_{\balpha_n}(\mr{\ell}) + e_{n,\ell}
          \vspace{-1mm}
\end{equation*}
with $\psi_{\balpha_n}$ an unknown nonlinear function to be estimated that defines the interaction between the endmember spectra, in the proportion $\balpha_n$, and $\be_n$ the estimation error. This leads us to consider the general problem
\vspace{-2mm}
\begin{equation}
      \label{eq:psi}
      \psi_{\balpha_n}^* = \mathop{\arg\min}_{\psi_{\balpha_n}}  \frac{1}{2}\|\psi_{\balpha_n}\|_\cp{H}^2
      +\frac{1}{2\mu}\,\ssum{\ell=1}{L} (r_{n,\ell} \,- \psi_{\balpha_n}(\mr{\ell}))^2
\end{equation}
with $\mu$ a positive parameter that controls the trade-off between structural error and misadjustment error. Clearly, this basic strategy may fail if the functionals $\psi_{\balpha_n}$ cannot be adequately and finitely parameterized. In~\cite{Chen2013-TSP}, we defined them by a linear trend parameterized by the abundance vector $\balpha_n$, combined with a nonlinear fluctuation function $\psi_n$, namely,
\begin{equation}
	\label{eq:map}
    	\psi_{\balpha_n}(\mr{\ell}) = \balpha_n^\top\mr{\ell}+\psi_n(\mr{\ell})
\end{equation}
where $\psi_n$ can be any real-valued function in a reproducing kernel Hilbert space $\cp{H}$, endowed with the reproducing kernel~$\kappa$ such that $\psi_n(\mr{\ell}) = \langle\psi_n, \kappa(\cdot, \mr{\ell})\rangle$. Indeed, kernel-based methods lead to efficient and accurate resolution for inverse problems of the form~\eqref{eq:psi} by exploiting the central idea of this research area, known as the \emph{kernel trick}. We proposed in~\cite{Chen2013-TSP} to conduct data unmixing \eqref{eq:psi}--\eqref{eq:map} by solving the following least-square support vector regression (LS-SVR) problem
\begin{equation}
	\label{eq:problem.algo1}
	\begin{split}
      	\balpha_n^*,\psi_n^*&=\mathop{\arg\min}_{\balpha_n,\psi_n} \frac{1}{2}\Big(\|\balpha_n\|^2
	+\|\psi_n\|_{\cp{H}}^2 + \frac{1}{\mu}\|\boldsymbol{e}_n\|^2\Big) \\
	&\text{subject to} \quad \balpha_n \succeq \cb{0} \quad \text{and} \quad \cb{1}^\top\balpha_n = 1
      	\end{split}
\end{equation}
where $\boldsymbol{e}_n$ is the $(L \times 1)$ misadjustment error vector with $\ell$-th entry $e_{n,\ell}= r_{n,\ell} - (\balpha_n^\top\mr{\ell}+\psi_{n}(\mr{\ell}))$ as defined in~\eqref{eq:psi}. It can be shown that problem~\eqref{eq:problem.algo1} is convex so that it can be solved exactly by the duality theory. This so-called K-Hype method was introduced in~\cite{Chen2013-TSP}.  Finally, considering all the pixels of the image to process, the modeling error to be minimized is expressed as
\begin{equation*}
	J_{\text{err}}(\A,\bpsi)=\frac{1}{2}\ssum{n=1}{N} \Big(\|\balpha_n\|^2+\|\psi_n\|_{\cp{H}}^2 + \frac{1}{\mu}\|\boldsymbol{e}_n\|^2\Big)
\end{equation*}
subject to the contraints in~\eqref{eq:constraints}. In this expression, $\A = [\balpha_1,\dots, \balpha_N]$ and $\cb{\psi} = \{\psi_n\in\cp{H}\!:n=1,\dots,N\}$.

In order to take spatial correlation between pixels into account, we shall use $\ell_1$-type regularizers of the form~\cite{Zymnis2007,Iordache2012} to promote piecewise constant transitions in the fractional abundance of each endmember among neighboring pixels. The regularization function is expressed as
\vspace{-2mm}
\begin{equation}
	\label{eq:reg.elem}
    	J_{\text{sp}}(\A) = \ssum{n=1}{N}\ssum{m\in\cp{N}(n)}{}\|\balpha_n-\balpha_m\|_1
	\vspace{-1mm}
\end{equation}
where $\|\,\|_1$ denotes the $\ell_1$ norm, and  $\cp{N}(n)$ the set of neighbors of the pixel $n$.  Without loss of generality, in this paper, we define the neighborhood of a pixel $n$ by taking the $4$ nearest pixels $n-1$ and $ n+1$ (row adjacency), $n-w$ and $n+w$ (column adjacency). In this case, let us define the $(N\times N)$ matrices $\HH_{_\leftarrow}$ and $\HH_{_\rightarrow}$ as the two linear operators that compute the difference between any abundance vector and its left-hand neighbor, and right-hand neighbor, respectively. Similarly, let $\HH_{_\uparrow}$ and $\HH_{_\downarrow}$ be the linear operators that compute that difference with the top neighbor and the down neighbor, respectively. With these notations, the regularization function~\eqref{eq:reg.elem} can be rewritten in matrix form as
\begin{equation*}
	J_{\text{sp}}(\A) = \|\A\HH\|_{1,1}
\end{equation*}
with $\HH$ the $(N\times 4N)$ matrix $\left(\HH_{_\leftarrow}\,\HH_{_\rightarrow}\,\HH_{_\uparrow}\,\HH_{_\downarrow}\right)$ and $\|\,\|_{1,1}$ the sum of the $\ell_1$-norms of the columns of a matrix. Unfortunately, while this regularization function is convex, it is non-smooth.

Now considering both the mismodeling error $J_{\text{err}}$ and the regularization term $J_{\text{sp}}$, the optimization problem becomes
\begin{align}
 %    	\begin{split}
     		\A^*\!,\bpsi^*\!\!=&\mathop{\arg\min}_{\A,\cb{\psi}} \ssum{n=1}{N} \frac{1}{2} \Big(\!\|\balpha_n\|^2\!+\!\|\psi_n\|_{\cp{H}}^2 
				\!+\!\frac{1}{\mu}\|\boldsymbol{e}_n\|^2 \!\Big) \!\!  + \! \eta\,\|\A\HH\|_{1,1}  \nonumber  \\
                	&\text{subject to} \quad \A \succeq 0  \quad \text{and} \quad \A^\top\cb{1}_R = \cb{1}_N         	\label{eq:opt.prob}
%	\end{split}
\end{align}
The constraints over $\A$ define a convex set $\cp{S}_A$. For ease of exposition, we will denote the constraints by $\A\in\cp{S}_A$.

\section{Solving the problem}
\vspace{-2mm}
Although the optimization problem~\eqref{eq:opt.prob} is convex, it cannot be solved easily because it combines a functional regression problem with a large-dimensional non-smooth regularization term. In order to overcome this, we rewrite~\eqref{eq:opt.prob} in the following equivalent form
\begin{align}
%     \begin{split}
     		&\min_{\A\in\cp{S}_A,\cb{\psi}} \, \ssum{n=1}{N} \frac{1}{2}\Big(\|\balpha_n\|^2+\|\psi_n\|_{\cp{H}}^2
				+\frac{1}{\mu}\|\boldsymbol{e}_n\|^2 \Big) +  \eta\,\|\UU\|_{1,1} \nonumber\\
                	&\quad \quad\text{subject to} \quad \VV = \A  \quad \text{and} \quad \UU = \VV\HH         \label{eq:opt.prob.2}
 %     \end{split}
\end{align}
where two new matrices $\UU$ and $\VV$, and two additional constraints, have been introduced. This variable-splitting approach was initially proposed in \cite{goldstein2009split}. The matrix $\UU$ will allow us to decouple the non-smooth $\ell_1$-norm regularizer from the constrained LS-SVR problem. The matrix $\VV$ will make the LS-SVR problem tractable by relaxing connections between pixels.

\begin{figure*}[!t]
%\footnotesize
\setcounter{equation}{13}
\begin{equation}
	\label{eq:dual.algo1}
	\begin{split}
      		\max_{\bbeta_n,\bgamma_n,\lambda_n} \mathcal{L}_n'(\bbeta_n,\bgamma_n,\lambda_n) & =
				-\frac{\rho}{2\zeta}
			\left(
			\begin{array}{c}
				\bbeta_n \\ \hline
 				\bgamma_n \\ \hline
				\lambda_n
			\end{array}\right)^{\!\!\!\top}
			\left(
			\begin{array}{c|c|c}
				\KK_\psi  & \MM & -\MM\cb{1}_R \\ \hline
 				\MM^\top & \bI & -\cb{1}_R \\ \hline
				- \cb{1}_R^\top\MM^\top & -\cb{1}_R^\top & R
			\end{array}\right)
			\left(
			\begin{array}{c}
				\bbeta_n \\ \hline
 				\bgamma_n \\ \hline
				\lambda_n
			\end{array}\right)
			+ \left(
			\begin{array}{c}
				\br_n- \rho\,\MM\,\cb{\xi}_n^{(k)} \\ \hline
 				-\rho\,\cb{\xi}_n^{(k)} \\ \hline
				\rho\, \cb{\xi}_n^{(k)\top}\,\cb{1}_R-1
			\end{array}\right)^{\!\!\!\top}
			\left(
			\begin{array}{c}
				\bbeta_n \\ \hline
 				\bgamma_n \\ \hline
				\lambda_n
			\end{array}\right)
			\\
      		& \text{subject to} \quad \bgamma_n \succeq \bf{0} \\
            	& \text{with}\quad \KK_\psi = \frac{1}{\zeta}\,(\KK + \mu\,\bI)+\MM\MM^\top  \; \text{and} \; \rho=\frac{\zeta}{1+\zeta}
	\end{split} 
\end{equation}

% Restore the current equation number.
\setcounter{equation}{8}
% IEEE uses as a separator
\vspace{-2mm}
\hrulefill
% The spacer can be tweaked to stop underfull vboxes.
\vspace{-5mm}
\end{figure*}

As studied in~\cite{goldstein2009split}, the split-Bregman iteration algorithm is an efficient method to deal with a broad class of $\ell_1$-regularized problems. By applying this framework to \eqref{eq:opt.prob}, the following iterative formulation is obtained
\begin{align}
%	\begin{split}
    	&\A^{(k+1)},\cb{\psi}^{(k+1)},\VV^{(k+1)},\UU^{(k+1)}   \nonumber\\
	=&\mathop{\arg\min}_{\A\in\cp{S}_A,\cb{\psi},\VV,\UU}  \sum_{n=1}^N \frac{1}{2}\Big(\|\balpha_n\|^2\!+\!\|\psi_n\|_{\cp{H}}^2
				\!+\!\frac{1}{\mu}\|\boldsymbol{e}_n\|^2 \Big) \!+\! \eta\|\UU\|_{1,1} \nonumber  \\
   	&+\frac{\zeta}{2}\|\A-\VV-\bD_1^{(k)}\|_F^2 +\frac{\zeta}{2}\|\UU-\VV\HH-\bD_2^{(k)}\|_F^2    \label{eq:split.original}
%	\end{split}
\end{align}
with
     \vspace{-5mm}
\begin{equation}
      	\label{eq:split.complement}
	\begin{split}
     	& \bD_1^{(k+1)} = \bD_1^{(k)} +(\VV^{(k+1)}-\A^{(k+1)}) \\
     	&\bD_2^{(k+1)} = \bD_2^{(k)} +(\VV^{(k+1)}\HH-\UU^{(k+1)})
	\end{split}
\end{equation}
where $\|\,\|_F^2$ denotes the matrix Frobenius norm, and $\zeta$ is a positive parameter. Because of how we have split the terms of the cost function, we can now perform the above minimization efficiently by iteratively minimizing with respect to $(\A,\cb{\psi})$, $\VV$ and $\UU$ separately. The three steps we have to perform are:

\noindent\textbf{Step 1} - Optimization with respect to $\A$ and $\cb{\psi}$: The optimization problem \eqref{eq:split.original} reduces to
\vspace{-2mm}
\begin{equation*}
          \begin{split}
    	\A^{(k+1)},\cb{\psi}^{(k+1)} &=
				\mathop{\arg\min}_{{\A\in\cp{S}_A,\cb{\psi}}} \ssum{n=1}{N} \frac{1}{2}\Big(\|\balpha_n\|^2 + \|\psi_n\|_{\cp{H}}^2+\frac{1}{\mu}\|\boldsymbol{e}_n\|^2  \\
				& \hspace{1.8cm}+ \zeta\|\balpha_n-\cb{\xi}_n^{(k)}\|^2\Big)
	\end{split}			
\end{equation*}
where $\cb{\xi}_n^{(k)} = \VV_n^{(k)}+\bD_{1,n}^{(k)}$. Here, $\VV_n$ and $\bD_{1,n}$ denote the $n$-th column of $\VV$ and $\bD_1$, respectively. It can be observed that this problem can be solved, independently, for each vector $\balpha_n$. This results from the use of the matrix $\VV$. Let us now solve the local optimization problem
\vspace{-2mm}
\begin{align}
     %    \begin{split}
     &\balpha_n^{(k+1)},\psi_n^{(k+1)} \nonumber \\
     =    &   		\mathop{\arg\min}_{\balpha_n,\psi_n,\cb{e}_n}\! \frac{1}{2}\Big(\|\balpha_n\|^2 \!+\! \|\psi_n\|_{\cp{H}}^2
				\!+\!\frac{1}{\mu}\ssum{\ell=1}{L} e_{n,\ell}^2 \! +\! \zeta\|\balpha_n-\cb{\xi}_n^{(k)}\|^2\Big) \nonumber \\
      		&\text{subject to} \quad e_{n,\ell} = r_{n,\ell}- ( \balpha_n^\top\mr{\ell}  + \psi_n(\mr{\ell})) \nonumber \\
                &\phantom{\text{subject to}}\quad \balpha_n \succeq 0   \quad \text{and} \quad \balpha_n^\top\cb{1}_R = 1  \label{eq:opt.an}
 %     \end{split}
 \vspace{-2mm}
\end{align}
By introducing the Lagrange multipliers $\beta_{n,\ell}$, $\gamma_{n,\ell}$ and $\lambda_n$, where the superscript $(k)$ of these variables has been omitted for simplicity of notation, the Lagrange function associated with~\eqref{eq:opt.an} is equal to
\begin{align}
%    \begin{split}
    \mathcal{L}_n = & \frac{1}{2}\Big(\|\balpha_n\|^2 + \|\psi_n\|_{\cp{H}}^2
				+\frac{1}{\mu}\ssum{\ell=1}{L}\, e_{n,\ell}^2 + \zeta\|\balpha_n-\cb{\xi}_n^{(k)}\|^2\Big) \nonumber\\
          &-\ssum{\ell=1}{L} \beta_\ell(e_{n,\ell}-r_{n,\ell}+ \balpha_n^\top\mr{\ell} +\psi_n(\mr{\ell}))  \nonumber\\
          &-\ssum{r=1}{R}\gamma_r\alpha_{n,r}+\lambda_n(\balpha_n^\top\cb{1}_R -1)           \label{eq:Lagrange.primal}
%    \end{split}
\end{align}
with $\gamma_{n,r}\geq 0$. The conditions for optimality of $\mathcal{L}_n$ are
\begin{equation}
	\label{eq:solution.step1}
	\left\{
    		\begin{array}{ll}
    			\balpha_n^* = \frac{1}{\zeta+1}\left(\sum_{\ell=1}^L \beta_{n,\ell}^*\,\mr{\ell} 
			+ \bgamma_n^* - \lambda_n^*\cb{1}+\zeta\cb{\xi}_n^{(k)} \right)\\
    			\psi_n^* = \sum_{\ell=1}^L \beta_{n,\ell}^*\,\kappa(\cdot, \,\mr{\ell}) \\		
                e_{n,\ell}^* = \mu\,\beta_{n,\ell}^*
    	\end{array}
	\right.
\end{equation}
where $\kappa$ denotes the reproducing kernel of $\cp{H}$. By substituting \eqref{eq:solution.step1} into~\eqref{eq:Lagrange.primal}, we get the dual problem \eqref{eq:dual.algo1} (see above), where $\KK$ is the Gram matrix defined as $[\KK]_{\ell p}=\kappa(\mr{\ell},\mr{p})$. The problem \eqref{eq:dual.algo1} is a convex quadratic programming problem with respect to the dual variables. Finally, provided that the optimal dual variables $\bbeta_n^*$, $\bgamma_n^*$ and $\lambda_n^*$ have been determined, the vector of fractional abundances is estimated by
\vspace{-1mm}
\begin{equation*}
     	\balpha_n^* = \mathsmaller{\frac{1}{\zeta+1}}\big(\MM^\top\bbeta_n^* + \bgamma_n^* - \lambda_n^*\cb{1}+\zeta\cb{\xi}_n^{(k)} \big) 
	\vspace{-1mm}
\end{equation*}
This process has to be repeated for $n=1,\ldots,N$  to get $\A^{(k+1)}$.

\noindent\textbf{Step 2} - Optimization with respect to $\VV$: The optimization problem~\eqref{eq:split.original} now reduces to
\setcounter{equation}{14}
\begin{equation}
  	\label{eq:opt.V}
	\begin{split}
  	\VV^{(k+1)} = \mathop{\arg\min}_{\VV} & \|\A^{(k+1)}-\VV-\bD_1^{(k)}\|_F^2 \\ & + \|\UU^{(k)}-\VV\HH-\bD_2^{(k)}\|_F^2
	\end{split}
\end{equation}
Equating to zero the gradient of this expression with respect to $\VV$ directly gives us the solution
\begin{equation*}
      	\label{eq:solution.V}
      	\begin{split}
        \VV^{(k+1)} = \Big(\!\A^{(k+1)}\!-\bD_1^{(k)}+(\UU^{(k)}\!-\!\bD_2^{(k)})\,\HH^\top\Big)(\bI+\HH\HH^\top)^{-1}
        \end{split}
\end{equation*}

\noindent\textbf{Step 3} - Optimization with respect to $\UU$:  The optimization problem~\eqref{eq:split.original} reduces to
\begin{equation}
	\label{eq:opt.U}
	\begin{split}
  	\UU^{(k+1)} = \mathop{\arg\min}_{\UU} \eta\|\UU\|_{1,1} \!+\!\frac{\zeta}{2}\|\UU\!-\!\VV^{(k+1)}\HH\!-\!\bD_2^{(k)}\|_F^2
	\end{split}
\end{equation}
Its solution is expressed via the well-known soft threshold function
\begin{equation}
   	\label{eq:solution.U}
   	\UU^{(k+1)} = \text{Thresh}\big(\VV^{(k+1)}\HH+\bD_2^{(k)}, {\eta}/{\zeta}\big)
\end{equation}
where $\text{Thresh}(\cdot,\tau)$ denotes the component-wise application of the soft threshold function defined as~\cite{Tibshirani1996} 
\begin{equation*}
	\text{Thresh}(x,\tau)=\text{sign}(x)\,\max(|x|-\tau,0)
\end{equation*}
Note that, as they are spatially invariant, the multiplications by $\HH$ in the above expressions can be efficiently performed with an FFT.

%\vspace{-3mm}
\section{Experimental Results}
\vspace{-2mm}
\label{sec:experiments}
\begin{table*}[!th]
    \caption{RMSE comparison with the synthetic data.}
    \vspace{-2mm}
    \label{tbl:RMSE.cmp}
    %\footnotesize
\begin{center}
\begin{tabular}{|c||c|c|c|c||c|c|}
\hline
 &\multicolumn{2}{c|}{DC1 }&\multicolumn{2}{c||}{DC2}& \multicolumn{2}{c|} {Comp. time (ms/pixel)}\\
\cline{2-7}
 & Bilinear & PNMM  &Bilinear & PNMM &\phantom{sss}IM1\phantom{sss}&IM2\\
 \hline
FCLS  & 0.1730$\pm$0.0092 & 0.1316$\pm$0.0052 & 0.1680$\pm$0.0265 & 0.1444$\pm$0.0098 &0.07&0.08\\
%\hline
NCLS  & 0.1351$\pm$0.0131& 0.1468$\pm$0.0071 &  0.0784$\pm$0.0076& 0.1378$\pm$0.0135 &0.06&0.07\\
spatial.-reg. FCLS &0.1729$\pm$0.0091 & 0.1311$\pm$0.0052& 0.1676$\pm$0.0263 & 0.1381$\pm$0.0074 & 0.91&1.00\\
spatial.-reg. NCLS &0.1159$\pm$0.0044 &0.1472$\pm$0.0069 & 0.0685$\pm$0.0053 & 0.1304$\pm$0.0097   &0.85&0.90\\
K-Hype & 0.0781$\pm$0.0050& 0.0895$\pm$0.0072 & 0.0755$\pm$0.0080 & 0.1107$\pm$0.0104  &5.7&6.0\\
NK-Hype & 0.0771$\pm$0.0054& 0.0873$\pm$0.0066&0.0919$\pm$0.0082& 0.1059$\pm$0.0096  &5.7&6.0\\
spatial.-reg. K-Hype (proposed) & 0.0444$\pm$0.0016&0.0480$\pm$0.0480& 0.0521$\pm$0.0033  & 0.0849$\pm$0.0042  & 56.5&68.8\\
spatial.-reg. NK-Hype (proposed) & 0.0493$\pm$0.0026& 0.0458$\pm$0.0042 &0.0647$\pm$0.0032  & 0.0773$\pm$0.0044 &55.1& 69.8\\
\hline 
\end{tabular}
\end{center}
\vspace{-5mm}
\end{table*}
 
  \begin{figure*}
      \begin{center}
\includegraphics[width=0.20\textwidth]{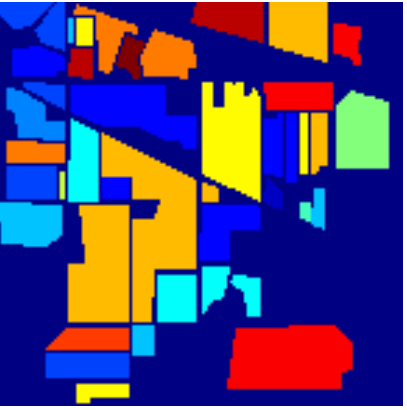} \qquad
\includegraphics[width=0.20\textwidth]{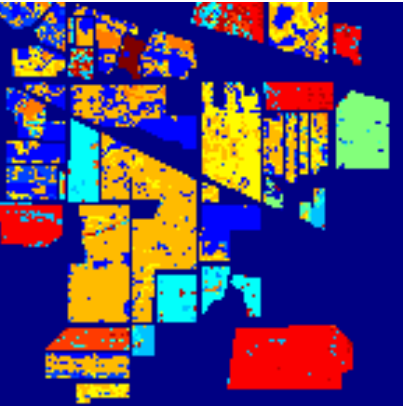} \qquad  
\includegraphics[width=0.20\textwidth]{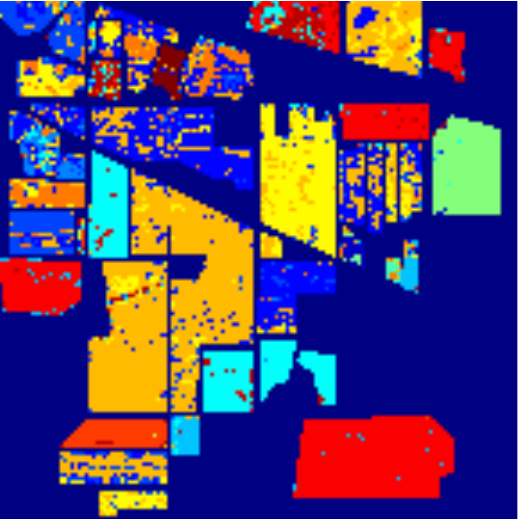}  \qquad
\includegraphics[width=0.20\textwidth]{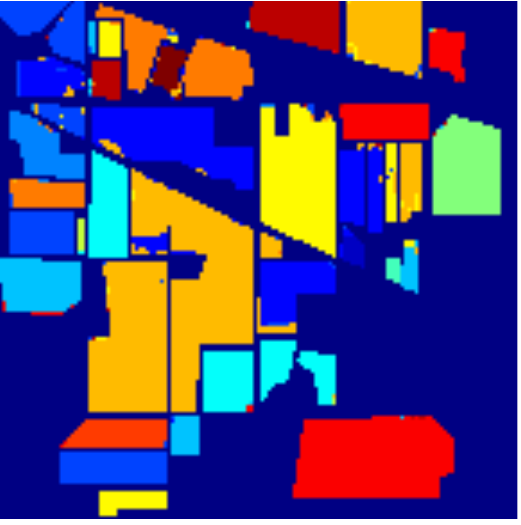} 
   \end{center}
   \vspace{-6mm}
       \caption{\label{fig:classification}\small Indian Pines classification map. From left to right: ground-truth, 
	FCLS (61.36\%), K-HYPE (71.39\%), Proposed (96.80\%).} 
	\vspace{-4mm}
\end{figure*}

\subsection{Experiments with synthetic images}
\vspace{-2mm}
Two spatially correlated hyperspectral images were generated for the following experiments. The endmembers were randomly selected from the spectral library ASTER~\cite{ASTER}, where signatures have reflectance values measured over $224$ spectral bands. Following~\cite{Iordache2012}, two spatially correlated abundance distributions, with $R=5$ and $R=9$ were used. See~\cite{Iordache2012} for the data description. The reflectance vectors were generated with two nonlinear mixture models described hereafter, and corrupted by a zero-mean white Gaussian noise $\bv_n$ with a SNR of $20$ dB. The first mixture model was the bilinear model defined as 
$      \br_n = \MM\balpha_n + \sum_{i=1}^{R}\sum_{j=i+1}^{R}\alpha_{n,i}\,\alpha_{n,j}\,\bm_i\otimes\bm_j + \bv_n,  $ 
with $\otimes$ the Hadamard product. The second one was a post-nonlinear model (PNMM) given by 
$         \br_n = (\MM\balpha_n)^{0.7} + \bv_n.    $
Several algorithms were tested in order to compare their unmixing performance on these two images. Their tuning parameters were set by preliminary experiments: 1)  The linear unmixing methods~\cite{Heinz2001}: The fully constrained least-square method (FCLS) was tested. By relaxing the sum-to-one constraint, one obtains the nonnegative constrained least-square method (NCLS), which was also considered. 2)  The spatially-regularized FCLS/NCLS: For comparison purposes,  regularizer~\eqref{eq:reg.elem} was considered with FCLS/NCLS algorithms, solved by split-Bregman iterations. 3)  The nonlinear unmixing algorithm K-Hype~\cite{Chen2013-TSP}: Unmixing was performed in this case by solving problem~\eqref{eq:problem.algo1}. Its nonnegative counterpart obtained by relaxing the sum-to-one constraint (NK-Hype) was also tested. The polynomial kernel defined by $\kappa(\mr{\ell},\mr{\ell}) =[1+(\mr{\ell}-1/2)^\top(\mr{\ell}-1/2)/R^2]^2$ was used, as in~\cite{Chen2013-TSP}.  4)  The proposed nonlinear algorithms incorporating spatial regularization: K-Hype and its nonnegative counterpart NK-Hype  were both considered with spatial regularization. The parameter $\zeta$ was adjusted in an adaptive way based on primal and dual residual norms at each iteration, see~\cite{Boyd2011}. Finally, the optimization algorithm was stopped when the number of iterations exceeded $10$, or both $\frac{\|\VV-\A\|_{F}}{N\times R}$ and $\frac{\|\UU-\VV\HH\|_{F}}{4 N\times R}$ became smaller than $10^{-5}$. \\ The RMSE 
\begin{equation}
	\text{RMSE} \!=\!\!{\sqrt{\mathsmaller{\frac{1}{NR}}\ssum{n=1}{N}\|\balpha_{n}-{\balpha^*_n}\|^2}}
\end{equation}
was used for comparing these algorithms, as reported in Table~\ref{tbl:RMSE.cmp}. Clearly, it can be observed that FCLS had large estimation errors. Relaxing the sum-to-one constraint with NCLS algorithm allowed to improve the performance in some cases, especially for DC2 with the bilinear model. The spatially-regularized FCLS and NCLS algorithms offered limited performance improvement. Nonlinear methods notably reduced this error in the mean sense, except for DC2 with the bilinear model. In this case, because most of the areas in the image are characterized by a dominant element with fractional abundance almost equal to one (see~\cite{Iordache2012} for visual illustration), mixing phenomena associated with the bilinear model are significantly weaker. Finally, the proposed spatially-regularized methods showed lower errors than all other tested algorithms.

\subsection{Experiments with AVIRIS data}
\vspace{-2mm}
In order to circumvent the difficulty that, in the literature, there is no available ground-truth for unmixing problems with real data, we adopted an indirect strategy to evaluate the proposed algorithm, via abundance-based classification. The estimated abundances were used as features to feed a classifier, and classification results were compared with labeled classification ground-truth. The scene used in our experiment is the well-known data set captured on the Indian Pines region by AVIRIS. The scene comprises $145 \times 145$ samples, consisting of $220$ contiguous spectral bands. The ground-truth data contains $16$ mutually exclusive classes.  This widely used benchmark data set is known to be dominated by mixed pixels, even if ground-truth information assigns each pixel to a unique class. In this experiment, the so-called unmixing based classification chain \#4 in~\cite{Dopido2011} was used. We tested FCLS, K-Hype, and the proposed algorithm for extracting abundance-based features. A one-against-all multi-class SVM with Gaussian kernel was applied to these data. We constructed five training sets by randomly selecting $5\%$, $10\%$, and $15\%$ of the samples available per class. All the required parameters were optimized by preliminary experiments. Table~\ref{tab:classification} summarizes the classification accuracies of SVM operating on features extracted with the unmixing algorithms. Fig.~\ref{fig:classification} presents these results in the case of an SVM trained with $10\%$ of the samples available per class. It appears that our nonlinear unmixing algorithms are more efficient than the linear one for feature extraction.  Finally, we observe that spatial regularization  greatly improved the classification accuracy. 
\begin{table}
 	\center
 	\caption{\label{tab:classification} \hspace{-1mm}Classification performance with abundance-based features.}
	\vspace{1mm}
	\begin{tabular}{|c|c|c|c|c|}
  		\hline
    		 		& $5\%$ 		& $10\%$		& $15\%$		\\
		\hline   
 		FCLS   	& $56.41$		& $61.36$		& $62.32$		\\
  		\hline        
  		K-Hype  	& $67.67$		& $71.39$		& $74.68$		\\
 		\hline
		Proposed	& $93.82$		& $96.80$		& $97.02$		\\
		\hline	
	\end{tabular}
	    \vspace{-5mm}
\end{table}

\section{Conclusion}
\vspace{-2mm}

We considered the problem of nonlinear unmixing of hyperspectral images. A nonlinear algorithm operating in reproducing kernel Hilbert spaces was proposed. Spatial information was incorporated using an $\ell_1$-norm local variation regularizer. Split-Bregman iterations were used to solve this convex non-smooth optimization problem. Experiments illustrated the effectiveness of this scheme.

% -------------------------------------------------------------------------
\newpage
%\vspace{-5mm}

\balance

\bibliographystyle{IEEEbib}
\bibliography{ref}
\vspace{-2mm}

\end{document}